\documentclass[10pt,twocolumn,letterpaper]{article}

\usepackage{xcolor}

\usepackage{wacv}
\usepackage{times}
\usepackage{epsfig}
\usepackage{graphicx}
\usepackage{amsmath}
\usepackage{amssymb}
\usepackage[caption=false, font=footnotesize]{subfig}
\usepackage[flushleft]{threeparttable}

\usepackage{enumitem}
\usepackage{pdfpages}
\usepackage{array, makecell} %
\usepackage{tabularx,booktabs}

\def\wacvPaperID{9} 

\wacvfinalcopy 

\ifwacvfinal
\def\assignedStartPage{1} 
\fi

\ifwacvfinal
\usepackage[breaklinks=true,bookmarks=false]{hyperref}
\else
\usepackage[pagebackref=true,breaklinks=true,colorlinks,bookmarks=false]{hyperref}
\fi

\ifwacvfinal
\else
\fi

\begin{document}

\title{Myope Models - Are face presentation attack detection models short-sighted?}

\author{\parbox{16cm}{\centering
    {\large Pedro C. Neto$^{1,2}$, Ana F. Sequeira $^{1}$ and Jaime~S.~Cardoso$^{2, 1}$}\\
    {\normalsize
    $^1$ INESC TEC, Porto, Portugal\\
    $^2$ Faculdade de Engenharia da Universidade do Porto, Porto, Portugal}}
   }

\maketitle

\begin{abstract}
Presentation attacks are recurrent threats to biometric systems, where impostors attempt to bypass these systems. Humans often use background information as contextual cues for their visual system. Yet, regarding face-based systems, the background is often discarded, since face presentation attack detection (PAD) models are mostly trained with face crops. This work presents a comparative study of face PAD models (including multi-task learning, adversarial training and dynamic frame selection) in two settings: with and without crops. The results show that the performance is consistently better when the background is present in the images. 
The proposed multi-task methodology beats the state-of-the-art results on the ROSE-Youtu dataset by a large margin with an equal error rate of 0.2$\%$. Furthermore, we analyze the models' predictions with Grad-CAM++ with the aim to investigate to what extent the models focus on background elements that are known to be useful for human inspection. From this analysis we can conclude that the background cues are not relevant across all the attacks. Thus, showing the capability of the model to leverage the background information only when necessary. 

\end{abstract}

\section{Introduction}

Presentation attacks are one of the weaknesses and dangers posed to face recognition systems (FRS). Apart from a few exceptions~\cite{kantarci2021shuffled}, most of the methods used in presentation attack detection (PAD) rely on tight face crops~\cite{kisku2017face}. In other words, cropping the faces removes everything in the image that is not part of a face. Thus, the information provided as input for these systems is somehow limited, leading to myopic (i.e. short-sighted) face PAD models. This pre-processing step has several advantages, including the capability to process several faces per frame and it is a fact that first generation face PAD methods were designed to rely only on the face region, in order to be backwards compatible with FRS. However, on the other hand, it removes spatial and contextual information present in the original image. The human visual cortex can process this spatial and contextual information to identify some attacks meant to fool the human inspection. Moreover, in some cases, a human can still be fooled by face crops of replay attacks, if the resolution of the replay device is high enough. Similarly, machine vision systems can likely learn to leverage that information when it is available. Furthermore, they can decide if it is more important to focus on the contextual information or on the face itself.

In this work, we aim at further studying an alternative approach to discarding upfront the background, already adopted in some previous works in the literature. Thus, in this work we investigate how different approaches perform in the presence of contextual background.
In the experiments, we perform a comparative study of a state-of-the-art supervised binary classification model and its combination with an adversarial approach (in this, two embeddings are produced, one containing information that is useful for the prediction and the other containing nuisances present in the input, which are optimized to minimize the mutual information between them), on three datasets, using face images with and without crops. For a more thorough evaluation, we elected the ROSE-Youtu dataset due to the fact that it offers a high variety of attacks not available in the other datasets. Thus, we evaluated a multi-task model which was optimized to also distinguish between the different attacks and the previous combined with an adversarial approach. Furthermore, we designed a novel experiment based on multiple instance learning methods. With this, we attempted at creating a dynamic frame selection system, passing the responsibility of selecting the frame most likely to include an attack to the model. Differently from the previous approaches, this method requires more processing power since it goes through all frames in a video.

It has been shown in the literature that black-box models, such as deep neural networks, can learn unpredictable patterns and focus their decision on ``unexpected'' regions of interest in the input. Therefore, we also evaluated the experiments from the perspective of explainable artificial intelligence (xAI). These evaluations are necessary to better understand the models' decisions and the errors due to their opaqueness~\cite{Hutson2021}. We use methods for the visualization of the elements that were important for the decision, such as  Grad-CAM++~\cite{Chattopadhay2018}. The output of these methods allowed us to analyze the visual cues found by the model to detect attacks. We also note that in some cases, the models follow the same cues used by an human inspector. Thus, we reflect upon the influence of the background in the choice of future face PAD algorithms.

This study contributes to the awareness around the need to incorporate interpretability in face PAD methodologies. The models' performance improvements are attributed to the use of background and this can be corroborated by observing that, in fact, the models' decisions are made using the background cues.


The experiments use three datasets: ROSE-Youtu~\cite{Haoliang2018}; NUAA~\cite{tan2010face}; and Replay-attack~\cite{chingovska2012effectiveness}. The state-of-the-art supervised binary classification model (BC), and BC combined with an adversarial approach are evaluated on the three datasets in two settings: with and without background. Furthermore, we evaluate the multi-task (MT) and dynamic frame selection (DFS) approaches using only the ROSE-Youtu dataset which includes a high diversity of attacks comprising both two-dimensional and three-dimensional information. Hence, it was used to study the impact of the background in the model performance and whether the background affects the capability of generalizing between attacks. However, the experiments were defined with two goals in mind: generalization between attacks, and generalization between subjects. The first goal is addressed, through the study of the performance of the model on attacks that were not seen previously during training. The second is addressed by using 50\% of the subjects for testing. The BC, MT and adversarial approaches are evaluated with and without background in a cross-dataset scenario. 

The major contributions of this work are: i) the evaluation (in three widely used datasets) of a state-of-the-art supervised binary classification model and its combination with an adversarial strategy in two alternative scenarios: face images with and without background; ii) a multi-task face PAD approach that leverages background and achieves state-of-the-art results on the ROSE-Youtu dataset; iii) a proposed methodology for frame selection strategy on the ROSE-Youtu dataset. It was not the focus of this study to see if specific models in the literature perform better with and without background, instead, it focuses on proposing simple and distinct approaches and analyzing whether the results are consistent across them with regard to the presence of background. 

Besides this introduction and the conclusion, this document contains four major sections. First, in Section~\ref{sec:related} there is a discussion of the related work and how it led to the current study. Afterwards, details on the experiments conducted are given in Section~\ref{sec:methods}. The description of the datasets is given in Section~\ref{sec:dataset}.  And finally, in Section~\ref{sec:results} we present and discuss the obtained results and how it impacts the future of PAD methods. 

\section{Related Work}
\label{sec:related}


Typically, previous works on face presentation attack detection do not leverage background information. And thus, removing it is a common practice and a frequent step on the preprocessing stage. The most common approach is to use a face crop, usually obtained through the use of deep learning-based face detection algorithms such as MTCNN~\cite{Zhang2016} and RetinaFace~\cite{Deng_2020_CVPR}. Earlier methods used more traditional techniques, for instance, the Viola-Jones cascade detector~\cite{viola2001}.


Within the published works, it is possible to find reinforcement learning approaches~\cite{cai2020drl}, 3D-CNNs~\cite{li2018}, a two stages approach relying on blinking~\cite{Hasan2019} and several other colour-based methods~\cite{Haoliang2018,Boulkenafet2016}. The background usage is addressed in some works~\cite{vareto2021face,baweja2020anomaly,nowara2017ppgsecure,yang2014learn,kantarci2021shuffled}, however, they did not perform comparative studies regarding performance, with and without the background, of several approaches. It is possible to find this comparison in other works, however, the proposed methods are based on conventional machine learning instead of end-to-end deep learning~\cite{yang2013face,komulainen2013context}. Due to the nature of the ROSE-Youtu dataset, which contains three-dimensional and two-dimensional attacks, there are fewer methods tested on this dataset than on others. The variability of attacks included in the dataset significantly increase the difficulty of finding a model capable of performing well on all of them. For this reason, even methods that achieve almost zero error on other datasets, have worse performance on the ROSE-Youtu~\cite{Du2021}.

To the best of our knowledge, there has not been any method inspired by multiple instance learning applied to face PAD. However, there is an article on a similar technique used for the detection of deep fakes~\cite{li2020sharp}. Despite being a slightly different problem, the detection methodology has a significant overlap. The adversarial approach followed in our experiments was described first at~\cite{NEURIPS2018Jaiswal} and its capabilities to work with face PAD systems was evaluated one year later by Jaiswal \textit{et al.}~\cite{jaiswal2019ropad}.

Producing and visualizing explanations of the predictions for face presentation attack detection is a relatively new topic. Sequeira \textit{et al.} have explored the challenge of interpreting face PAD methods~\cite{seq2020,Sequeira2021b}. Their work described how the current evaluation metrics for PAD lack information regarding the elements that are being used for the prediction. In a sense, they argue that models can make accurate predictions but still base their decision on parts of the image that do not correspond to real face features or presentation artifacts as a human inspector would. In this work, we follow a similar approach to produce and analyze explanations. However, we use them to infer if the presence of contextual background leads to the use of certain visual cues in the image. At the same time, we look forward to seeing if other contextual elements are used to make correct predictions, for instance, reflections. Humans often use these elements to make their analysis. 


\section{Methodology}
\label{sec:methods}

Myriads of attacks are constantly threatening biometric systems. However, in practice, we do not aim to identify the type of attack, thus the main goal is to infer if the image given to the sensor is an attack or if it is from a genuine person. The problem is, in its essence, formulated as a binary classification task. On top of the binary task, we also applied some different training processes. However, these do not affect the network at test time. The purpose of these distinct approaches is to understand if the background effect generalizes between approaches.

\begin{figure}[!h]
\centering
\includegraphics[width=2.5in]{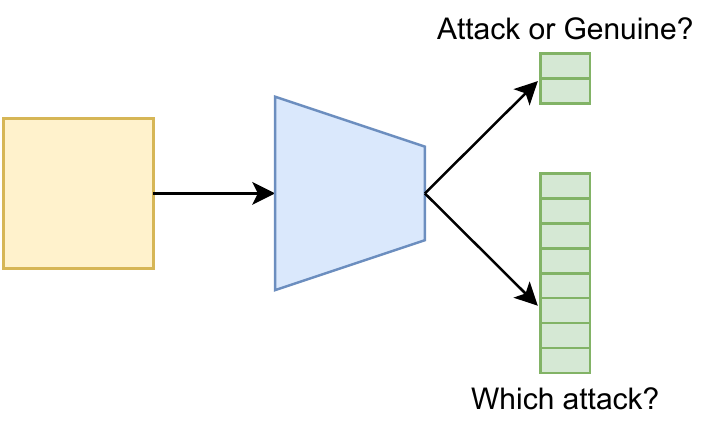}
\caption{Architecture of the multi-task learning model. It receives an image (yellow box) and includes a CNN (blue figure), two output heads (green figures), where the first is a binary head and the second has 8 output nodes. The output node for the genuine samples on both heads ensures that the main goal remains on the detection of attacks. }
\label{multi_task_learning}
\end{figure}

\begin{figure}[!h]
\centering
\includegraphics[width=2.7in]{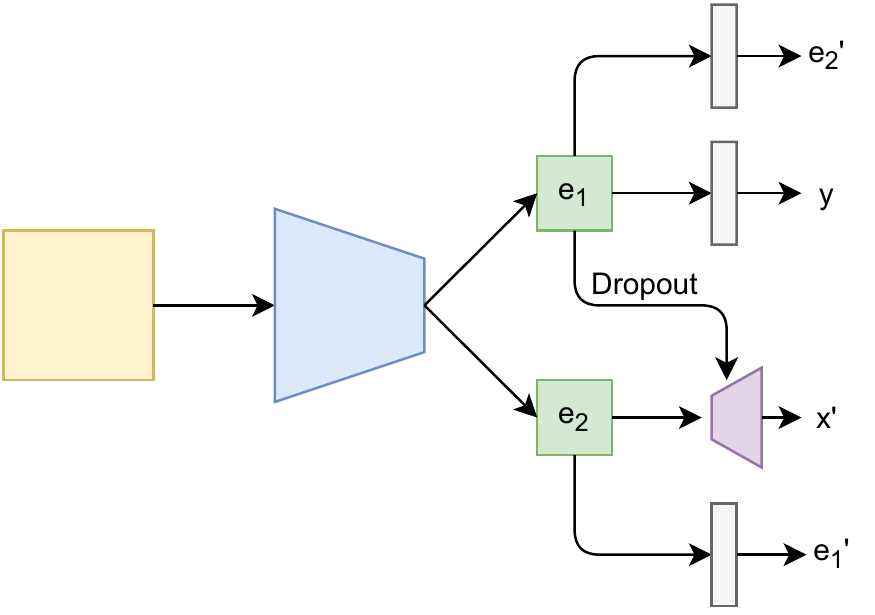}
\caption{Architecture of the adversarial learning model. It receives an input image (yellow box) and includes a CNN (blue figure), two feature vectors e\textsubscript{1} and e\textsubscript{2} (green boxes). These are used to reconstruct each other, decode (purple figure) the input and to classify the input. This architecture is deeply based on the Unsupervised Adversarial Invariance~\cite{NEURIPS2018Jaiswal}.}
\label{adversarial_architecture}
\end{figure}

\begin{itemize}[leftmargin=*]
    \item \textit{Binary classification training (BC)} - In the first approach, the only task that the backbone network is optimized for at training time is to classify between attacks and genuine. For this, we use a MobileNet v2~\cite{sandler2018mobilenetv2} that outputs two values, which are activated with softmax. The optimization of the weights is done using the binary cross-entropy loss (Eq.~\ref{eq:bce}).
    
    \begin{equation}
        BCE(y,p) = -{(y\log(p) + (1 - y)\log(1 - p))}
    \label{eq:bce}
    \end{equation}
    
    \item \textit{Multi-task classification training (MT)} - Whenever a model is optimized to distinguish between attacks and genuine images, it treats all the attacks equally. However, in practice, the attacks are not the same, and each possesses distinctive characteristics. And thus, we also formulated the training stage of a MobileNet v2, so it learns to distinguish between the seven different attacks. It is also possible that learning to discriminate between attacks also boosts the performance whenever the attack is unknown. Instead of having an output layer with two classes, the network has two output layers. The first has two output classes, whereas the other has eight (seven attacks and one genuine). In both cases, they are activated with softmax. Both layers are, simultaneously, updated with the binary cross-entropy~\ref{eq:bce} and cross-entropy~\ref{eq:ce} losses, respectively. These losses are combined as seen on Equation~\ref{eq:mt}. Due to the risk of the second term of the equation being larger than the first, it was necessary to add an output node for genuine samples in both heads.
    Figure \ref{multi_task_learning} is a simplified visualization of the architecture of the model. 
    
    \begin{equation}
        CE(y,p) = -\sum_{c=1}^My_{o,c}\log(p_{o,c})
    \label{eq:ce}
    \end{equation}
    
    \begin{equation}
        Loss_{Multi}(y_{1},p_{1},y_{2},p_{2}) = BCE(y_{1},p_{1}) + CE(y_{2},p_{2})
    \label{eq:mt}
    \end{equation}
    
    \item \textit{Adversarial training (Adv.)} - In the images shown to the system, there is background information that is useful for the prediction, for instance, reflections. Nevertheless, not all the background information is useful. Part of it can be considered to be a nuisance. Hence, we explored also an approach that attempted to remove those parts of the image from the feature vector used for the classification task. This approach, known as Unsupervised Adversarial Invariance~\cite{NEURIPS2018Jaiswal}, produces two distinct embeddings (i.e. feature vectors). The first vector, e\textsubscript{1}, represents the features that are relevant to the prediction of the model, whereas the second, e\textsubscript{2}, comprises the information that should not be used for the prediction. Constructing the loss of such architecture requires four terms. The first two are maximization terms (Eq.~\ref{eq:main_} and ~\ref{eq:main_mt}). They attempt to reconstruct e\textsubscript{1} from e\textsubscript{2} and vice-versa. This attempts at removing any potential mutual information between both embeddings. The other two are minimization terms (Eq.~\ref{eq:adver}). The first embedding uses e\textsubscript{1} to perform the classification task. Whereas the second apply some noise to e\textsubscript{1}, in the form of a dropout layer. From the noisy e\textsubscript{1} and from e\textsubscript{2}, it tries to reconstruct the input image. For the construction/reconstruction terms the loss used is the mean squared error (Eq.~\ref{eq:mse}), while for the classification term we use either the binary cross-entropy (Eq.~\ref{eq:bce}) or the multi-task loss (Eq.~\ref{eq:mt}). The term $\alpha$ controls the impact of the reconstruction loss on the overall loss. We start with $\alpha=0.025$, and we increase by 0.025 at the end of each epoch. The architecture is represented in Figure~\ref{adversarial_architecture}.
    
    \begin{equation}
        MSE(x,y) = \sum_{i=1}^{D}(x_i-y_i)^2
    \label{eq:mse}
    \end{equation}
    
    \begin{equation}
        Loss_{Adv}(e_1,e_2,e_1',e_2') = -MSE(e_1,e_2') - MSE(e_2,e_1')
    \label{eq:adver}
    \end{equation}
    
    \begin{equation}
\begin{split}
Loss_{Class}(y,p,x,x')& = BCE(y,p)  + \alpha MSE(x,x') 
\end{split}
\label{eq:main_}
\end{equation}

\begin{equation}
\begin{split}
Loss_{Class}(&y_{1},p_{1},y_{2},p_{2},x,x') = \\
 &\quad Loss_{Multi\_task}(y_{1},p_{1},y_{2},p_{2}) \\
 &\quad + \alpha MSE(x,x') 
\end{split}
\label{eq:main_mt}
\end{equation}

\end{itemize}

\begin{figure}[!h]
\centering
\includegraphics[width=2.5in]{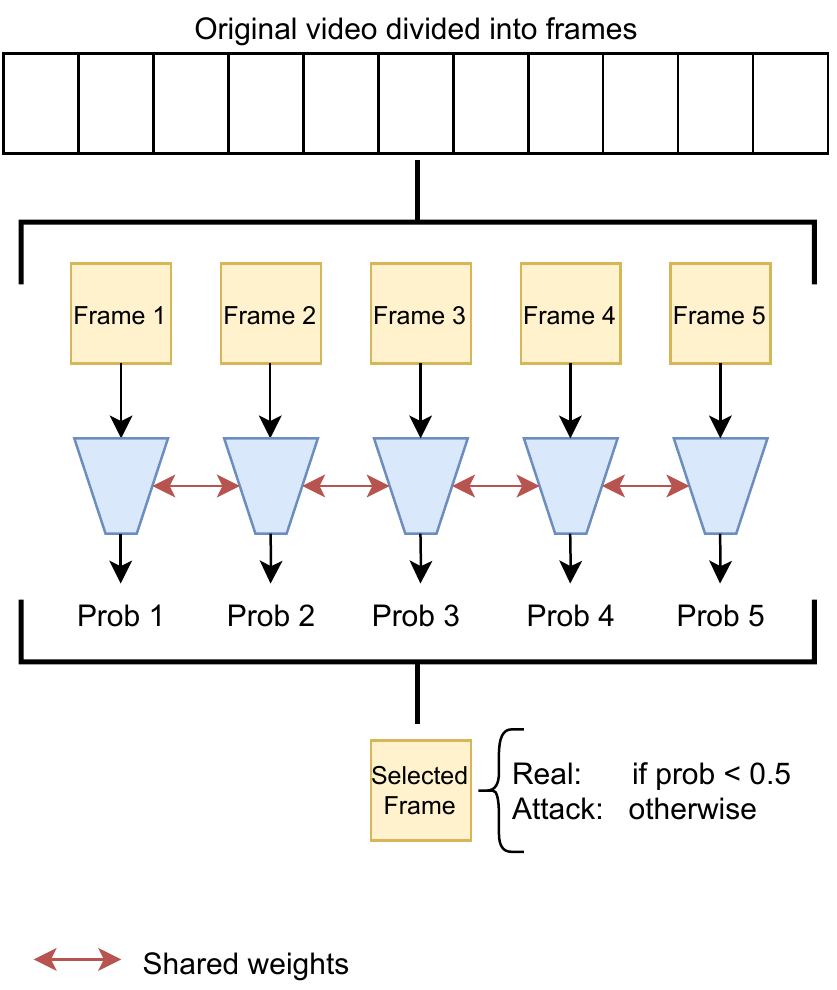}
\caption{Architecture of the method for dynamic frame selection. It includes a CNN (blue figure) with shared weights that processes all the frames (yellow boxes) of a video and selects one (lower yellow box) based on a specific criteria.}
\label{mil_architecture}
\end{figure}

\begin{itemize}[leftmargin=*]
    \item \textit{Dynamic frame selection training (DFS)} - Finally, we also propose an architecture to select the best frame for the detection of attacks. In state-of-the-art approaches, the training frames were fixed and previously selected from the list of possible frames. This, however, raises a couple of questions: 1) Do all the frames contain the same information for the prediction?; 2) If not, are we selecting the best frames to optimize the network?. We structured the optimization of this method in two stages: frame selection and learning. The frame selection stage processes all the frames in a video and computes their output with the model. From the outputs, if the video is from an attack it selects the three frames that have the lowest probability of being an attack. And the opposite if the video is from a genuine individual. Afterwards, the selected frames are used in the learning stage to optimize the network towards the video labels. At testing time the process is similar to the frame selection, the frame with the highest probability of being an attack is used for the classification. Perhaps, one of the most interesting aspects of this approach is that it can be integrated with the other previously mentioned. The frame selection step remains unchanged, while the training stage integrates the changes related to the other approaches.  Figure~\ref{mil_architecture} shows the behavior of the described method for both frame selection and testing. The frame selection probability is the result of the attack probability produced by the binary classification layer of the model.
\end{itemize}

To evaluate and compare the performance of these PAD models, we collected the following metrics: the \textit{Bona fide Presentation Classification Error Rate (BPCER)} (the proportion of bona fide presentations erroneously classified as attacks), and the \textit{Attack Presentation Classification Error Rate (APCER)} (the proportion of presentation attack wrongly classified as bona fide)~\cite{ISOPAD2017}. Finally, we also collected the \textit{Equal Error Rate (EER)}, which is the error at the operation point where the APCER and BPCER have the same value. For the APCER and the BPCER we used a threshold of 0.5.

\section{Datasets}
\label{sec:dataset}

The datasets used for the experimental evaluation are: ROSE-Youtu~\cite{Haoliang2018}; NUAA~\cite{tan2010face}; and Replay-Attack~\cite{chingovska2012effectiveness}.

\textbf{ROSE-Youtu~\cite{Haoliang2018}}: Contains, in its public version, 3350 videos with 20 different subjects. On average, video clips have a duration of 10 seconds. For each of the subjects, it contains around 150 to 200 videos captured from five mobile devices (all with different resolutions on their camera) and five lighting conditions. The front-facing camera was used with a distance between face and camera of about 30 to 50 centimeters.

\begin{figure}[h!]
    \centering
    \subfloat[Cropped \\Attack \#4 \label{4_crop}]{%
       \includegraphics[height=2.1cm,width=1.5cm]{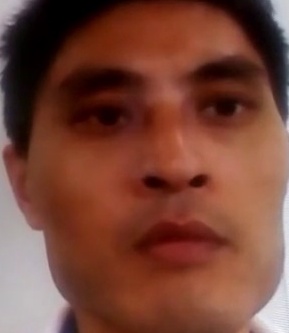}}
    ~
    \subfloat[Cropped \\Attack \#1 \label{1_crop}]{%
        \includegraphics[height=2.1cm,width=1.5cm]{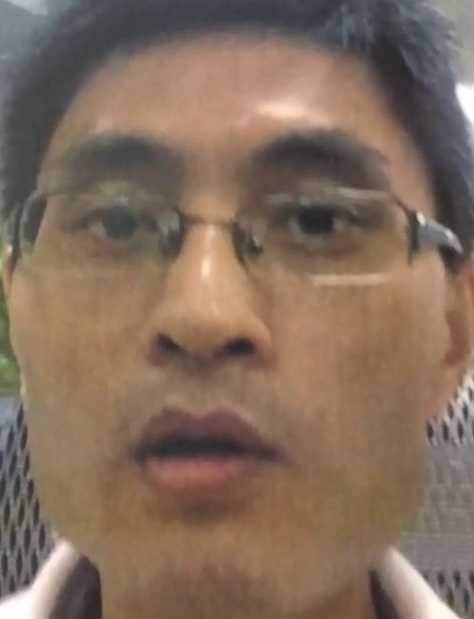}}
    ~
    \subfloat[Cropped \\Attack \#6\label{6_crop}]{%
        \includegraphics[height=2.1cm,width=1.5cm]{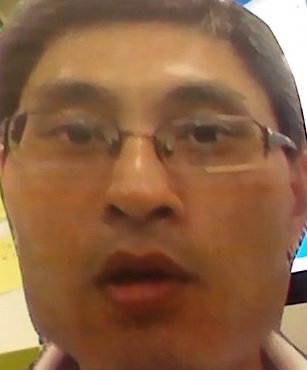}}
    ~
    \subfloat[Cropped \\Genuine\label{genuine_crop_1}]{%
        \includegraphics[height=2.1cm,width=1.5cm]{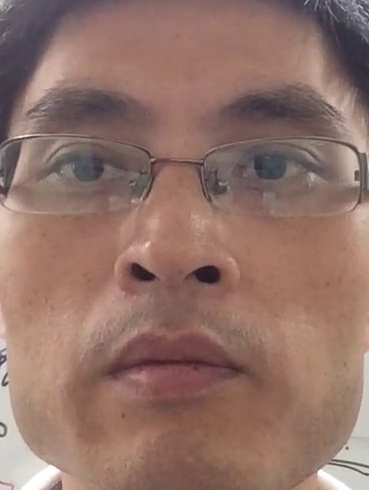}}
    ~
    \subfloat[Cropped \\Genuine\label{genuine_crop_2}]{%
        \includegraphics[height=2.1cm,width=1.5cm]{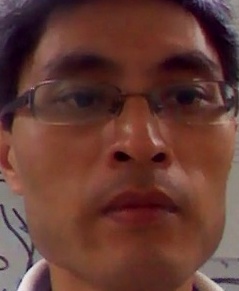}}
    \\
    \subfloat[Attack \#4 \label{4}]{%
       \includegraphics[height=2.1cm,width=1.5cm]{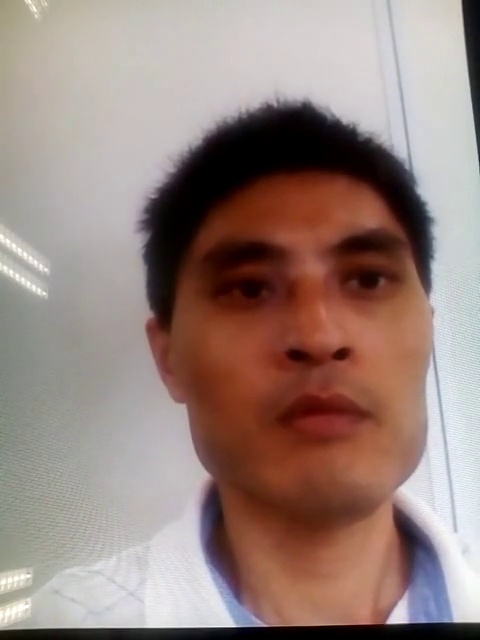}}
    ~
    \subfloat[Attack \#1 \label{1}]{%
        \includegraphics[height=2.1cm,width=1.5cm]{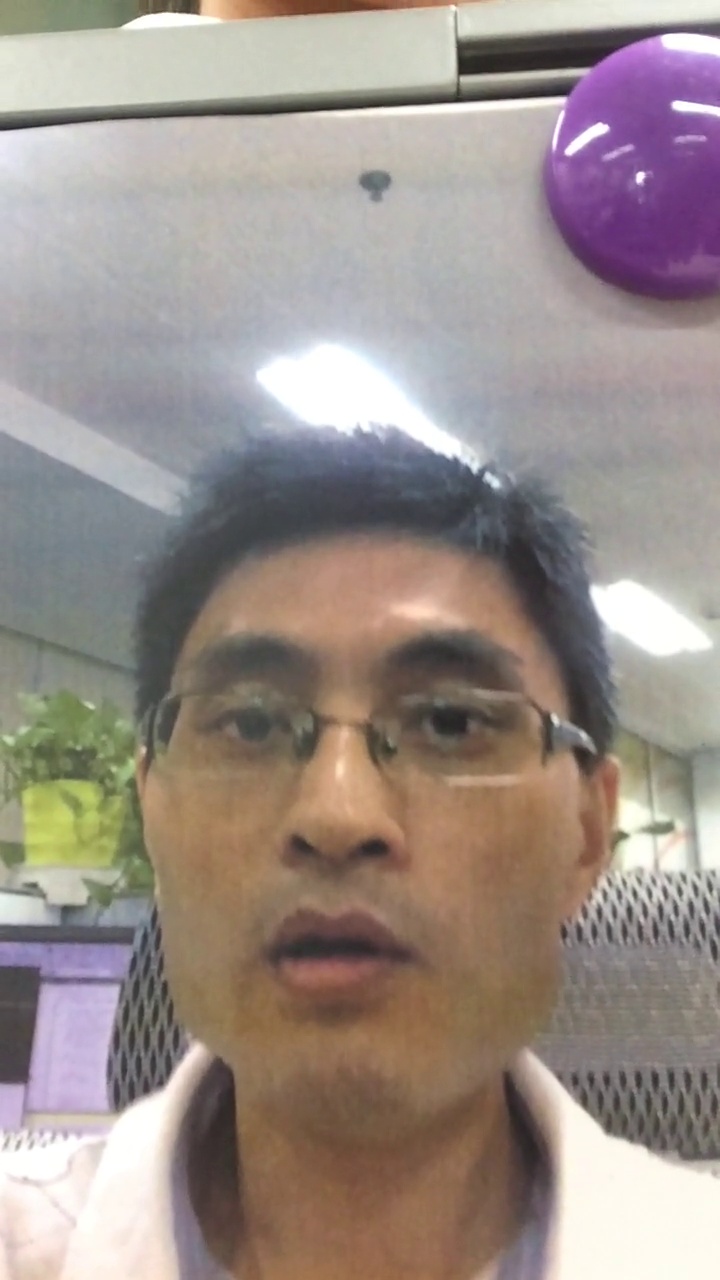}}
    ~
    \subfloat[Attack \#6 \label{6}]{%
        \includegraphics[height=2.1cm,width=1.5cm]{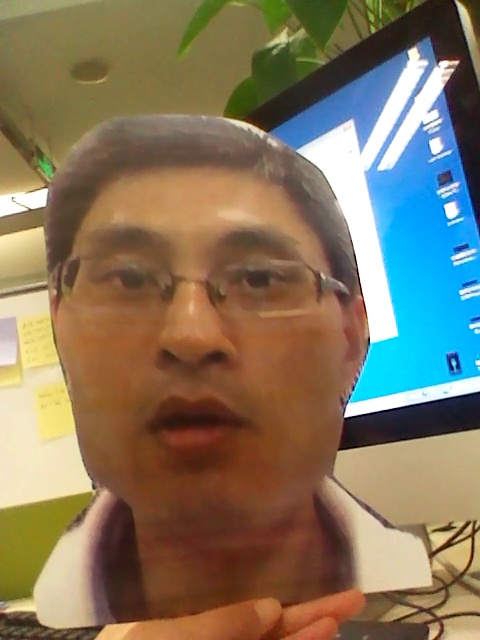}}
    ~
    \subfloat[Genuine\label{genuine_1}]{%
        \includegraphics[height=2.1cm,width=1.5cm]{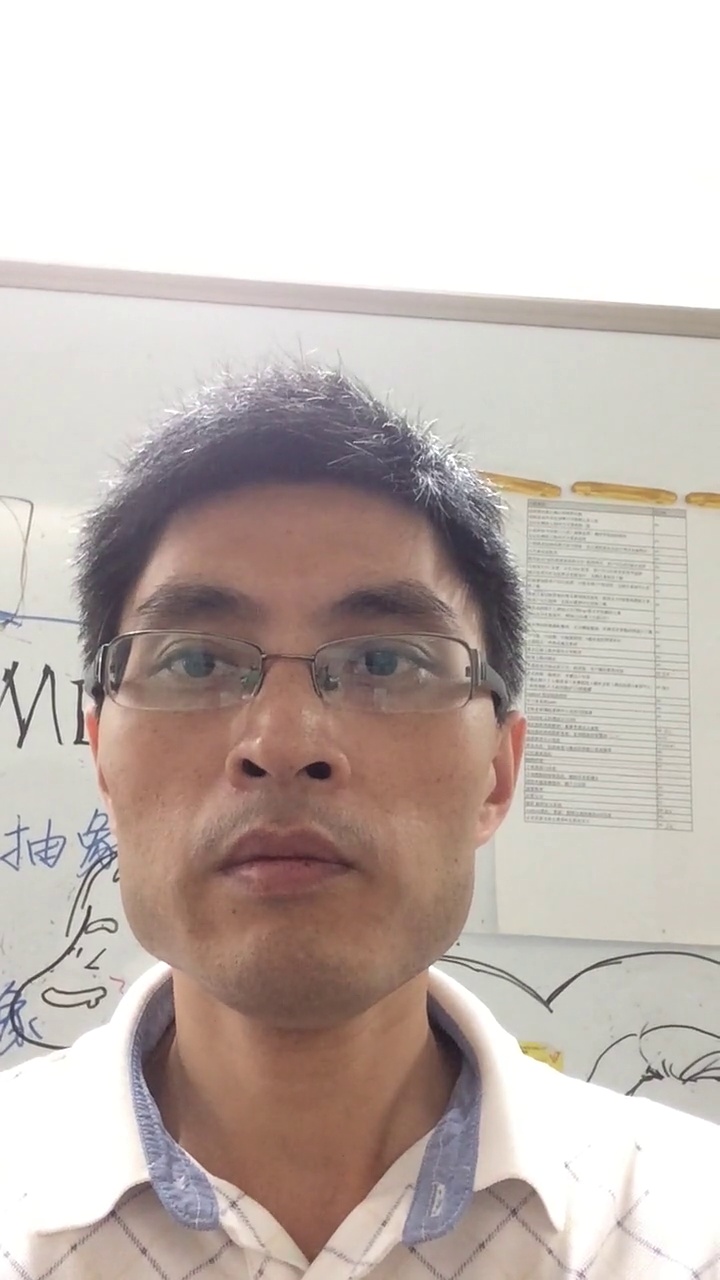}}
    ~
    \subfloat[Genuine\label{genuine_2}]{%
        \includegraphics[height=2.1cm,width=1.5cm]{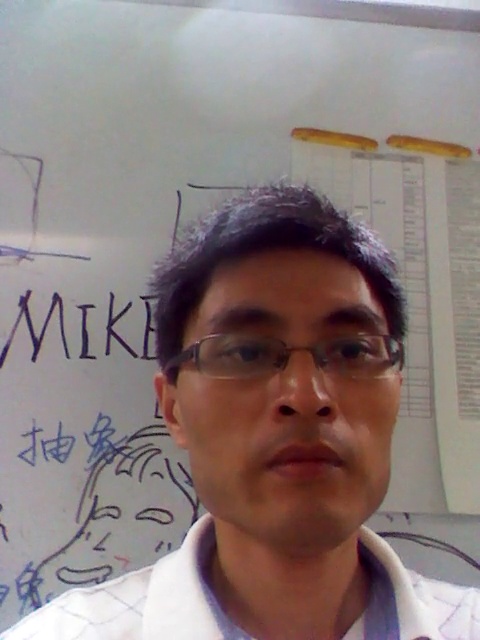}}

    \caption{Samples collected from the ROSE-YOUTU dataset~\cite{Haoliang2018} containing images from attacks and genuine captures. On the top row, cropped images are displayed. Whereas the bottom row contains the exact same images, but with all the background information included.}
    \label{cropped_attacks}
\end{figure}

\begin{table}[!h]
 \caption{List of attacks present in the ROSE-YOUTU dataset~\cite{Haoliang2018}.}
\label{table_attacks}
\centering
\begin{tabular}{lc}
\toprule
Attack     &  Description \\ 
\midrule
  - & Genuine (bona fide) \\
 \#1 & Still printed paper  \\
 \#2 & Quivering printed paper  \\ 
 \#3 & Video which records a Lenovo LCD display \\ 
 \#4 & Video which records a Mac LCD display  \\ 
 \#5 & Paper mask with two eyes and mouth cropped out \\ 
 \#6 & Paper mask without cropping \\ 
 \#7 & Paper mask with the upper part cut in the middle \\ 

\bottomrule
\end{tabular}
\end{table}

There are eight different types of videos, which translates into eight classes. The first class represents the genuine samples, whereas each of the following seven represent an attack. The first two attacks are print attacks, while the third and fourth are replay attacks on a Lenovo and an Apple laptop, respectively. The remaining three are based on paper masks and are responsible for including three-dimensional information in the dataset. These attacks are described in Table~\ref{table_attacks}.

We preprocessed the dataset into two different copies. For both, the frames of the videos were extracted and stored. For the second, a face was cropped by the MTCNN algorithm~\cite{Zhang2016} for all frames.  Examples of these images are seen in Figure~\ref{cropped_attacks}.
We used the videos from the first 10 indexed subjects (2,3,4,5,6,7,9,10,11,12) for training and the remaining 10 for testing.

\textbf{NUAA~\cite{tan2010face}}: was one of the first public databases for training and evaluating the performance of face PAD methods. This database simulates a simple and general method that re-captures a printed photograph of users for attacking a face recognition system. The NUAA database contains real and presentation attack face images of 15 persons. For each person, both real and presentation attack images were captured in three different sessions using generic cheap webcams and real face and printed photograph of users. The NUAA database contains 5105 real and 7509 presentation attack face images in color space with 640 × 480 pixels of image resolution. In this database, using the collected images, the training and testing sub-databases are predefined for training and testing of the PAD method, through which the performances of various PAD methods can be compared. In detail, the training database contains 1743 real and 1748 presentation attack face images, while the testing database contains 3362 real and 5761 presentation attack face images.

\textbf{Replay-Attack~\cite{chingovska2012effectiveness}}: this database for face PAD consists of 1300 video clips of photo and video attack attempts to 50 clients, under different lighting conditions. All videos are generated by either having a (real) client trying to access a laptop through a built-in webcam or by displaying a photo or a video recording of the same client for at least 9 seconds.

\section{Results}
\label{sec:results}

In this section, we present and discuss the evaluation results of the methods described. Regarding implementation details, all the methods described were optimized with Adam and a fixed learning rate of 0.001. The model used is a MobileNet v2 pre-trained on the ImageNet dataset. The images are resized to have a resolution of 224x224 and an RGB color scheme. The Grad-CAM++~\cite{Chattopadhay2018} was the visualization tool used to analyze the parts of the image relevant to the models' decisions. Train and test set splits were the same described in the publication of each dataset, so that the results can be compared with other works.

In Table~\ref{results_train_other} are shown the results obtained with the binary classification (BC) and its combination with the adversarial training (Adv.+BC) for the NUAA and Replay-Attack datasets. From both evaluations it is evident the performance improvement when the background is present in the images with a decrease in the EER to $0.00\%$.

The variety of attacks present in the ROSE-Youtu dataset motivates a multi-task learning approach. The experiments produced intended to evaluate the BC and the MT approaches with the inclusion and exclusion of contextual background. We further attempted to integrate them with the adversarial approach described in the previous section. The results for both BC and MT classification and their combination with the adversarial training strategy (Adv.+BC; Adv.+MT) can be seen in Table~\ref{results-no-back}. While the adversarial training did not lead to the expected improvement, in all four scenarios the models' performance improved with the background. This seems to indicate that the background provides more information and favoured the performance error rates. On multi-task classification, the improvements on the EER are as high as 81\%.


\begin{table}[ht!]
 \caption{Comparison of four different approaches with their versions with and without background. The columns represent the dataset used for both training and testing. The reported values represent the EER in \%.  
 }
\label{results_train_other}
\centering
\begin{tabular}{lccc}
\toprule
Method & Background &  NUAA & Replay-Attack    \\ 
\midrule
 BC  & \makecell{No\\Yes} & \makecell{2.91\\\textbf{0.00}} &  \makecell{\textbf{0.00}\\\textbf{0.00}}   \\
 \cline{1-4}
 Adv.+BC & \makecell{No\\Yes} & \makecell{3.03\\\textbf{0.00}} &  \makecell{0.33\\\textbf{0.00}}   \\

\bottomrule
\end{tabular}
\end{table}

\begin{table}[ht!]
 \caption{Comparison of four different approaches with their versions with and without background on the ROSE-Youtu dataset. APCER, BPCER and EER are displayed as \%.  In bold is the best result per column. }
\label{results-no-back}
\centering
\begin{tabular}{lcccc}
\toprule
Method & Background &  APCER & BPCER & EER  \\ 
\midrule
 BC  & \makecell{No\\Yes} & \makecell{0.49\\\textbf{0.25}} &  \makecell{2.20\\\textbf{2.03}}  & \makecell{1.32 \\\textbf{0.73}}  \\
 \cline{1-5}
 MT  & \makecell{No\\Yes}& \makecell{1.34\\\textbf{0.15}} &  \makecell{1.17 \\\textbf{0.40}}  & \makecell{1.26 \\\textbf{0.24}} \\
 \cline{1-5}
 Adv.+BC & \makecell{No\\Yes} & \makecell{1.42\\\textbf{0.52}} &  \makecell{2.71\\\textbf{1.29}}  & \makecell{1.76\\\textbf{0.76}} \\
 \cline{1-5}
 Adv.+MT & \makecell{No\\Yes} & \makecell{1.18\\\textbf{0.29}} &  \makecell{2.93\\\textbf{1.11}}  & \makecell{1.91\\\textbf{0.60}} \\

\bottomrule
\end{tabular}
\end{table}

In a cross-dataset approach, we performed experiments in which the models trained with the ROSE-Youtu dataset were evaluated with the other two datasets. The results of these experiments are presented in Table~\ref{results-test-others}. These results are in line with the results depicted on Tables~\ref{results_train_other} and~\ref{results-no-back}. The comparison of the same-database and cross-database results show that the models' performance consistently improve when the background information is used.

\begin{table}[ht!]
 \caption{Comparison of four different approaches with their versions with and without background. Results for models trained on the ROSE-Youtu dataset and tested on the datasets of each column. The reported values represent the EER in \%. }
\label{results-test-others}
\centering
\begin{tabular}{lccc}
\toprule
Method & Background &  NUAA & Replay-Attack   \\ 
\midrule
 BC  & \makecell{No\\Yes} & \makecell{22.04\\\textbf{13.45}} &  \makecell{29.43\\\textbf{12.29}}   \\
 \cline{1-4}
 MT  & \makecell{No\\Yes}& \makecell{23.61\\\textbf{3.89}}&  \makecell{26.13\\\textbf{13.91}}  \\
 \cline{1-4}
 Adv.+BC & \makecell{No\\Yes} & \makecell{28.31\\\textbf{18.11}} &  \makecell{26.03\\\textbf{17.12}}  \\
 \cline{1-4}
 Adv.+MT & \makecell{No\\Yes} & \makecell{35.66\\\textbf{23.85}} &  \makecell{26.15\\\textbf{19.46}}  \\

\bottomrule
\end{tabular}
\end{table}

Considering the performance gains from the use of background, we used this scenario to explore the proposed multi-task (MT) and dynamic frame selection (DFS) strategies. The results of all approaches evaluated for the ROSE-Youtu dataset are presented in Table~\ref{comparison-approaches}. Unexpectedly, the performance of the DFS methods and the ones that used adversarial training produced worse results than the simple binary and multi-task classification. The BC and MT approaches performed well at detecting attacks, as can be seen by the low value of the APCER, $0.25\%$ and $0.15\%$, respectively. Regarding the detection of bonafide images, the BC performed worse than several of the other methods and the MT had the lowest BPCER of them all, $0.40\%$. 

\begin{table}[h!]
 \caption{Comparison of all the seven different approaches explored in the ROSE-Youtu dataset. All of the approaches leveraged background information. In bold is the best result per column. }
\label{comparison-approaches}
\centering
\begin{tabular}{lcccc}
\toprule
Method  &  APCER (\%) & BPCER (\%) & EER (\%) \\ 
\midrule
 BC  & 0.25 &  2.03 & 0.73  \\
 MT   &\textbf{0.15} &  \textbf{0.40} & \textbf{0.24}  \\
 Adv. + BC & 0.52 &  1.29 & 0.76  \\
 Adv. + MT & 0.29 &  1.11  & 0.60  \\
 DFS& 0.54 &  4.68  & 1.62  \\
 MT + DFS & 0.31 &  2.23  & 0.69  \\
 Adv. + DFS & 2.15 &  1.78  & 1.78  \\

\bottomrule
\end{tabular}
\end{table}

We extended the experiments of the multi-task classification approach for different evaluation configurations, one and unseen attack (the results can be seen in Table~\ref{one_attack} and Table~\ref{unseen_attack}, respectively). The multi-task classification was also integrated with the adversarial training and the dynamic frame selection for a better and more complete comparison. 

The one attack configuration selects one attack to be used for both training and testing. This is intended to see how hard is to overfit the model to that attack and to distinguish it from genuine images.  The results for this configuration are visible in Table~\ref{one_attack} and it is possible to see that while the three approaches are capable of overfitting to the majority of the attacks, the attack \#4 remains challenging to the adversarial and DFS approaches. 

\begin{table*}[h!]
 \caption{Evaluation of three approaches in the setting of one attack in the ROSE-Youtu dataset. In this setting, the attack in the first column is the only one used for training and testing. APCER, BPCER and EER are displayed as \%. In bold is the best result per column.}
\label{one_attack}
\centering
\begin{tabular}{lccc|ccc|ccc}
\toprule
Attack     & \multicolumn{3}{c}{MT} & \multicolumn{3}{c}{Adversarial MT} & \multicolumn{3}{c}{DFS MT}  \\ 
\cmidrule{2-10}
      &APCER &BPCER  &EER &APCER &BPCER &EER &APCER &BPCER &EER \tnote{c} \\
\midrule
  \#1 & \textbf{0.00}  &  \textbf{0.20}  &\textbf{ 0.05} &  0.20  &  0.29  & 0.27   &  0.50  &  0.22  & 0.50    \\ 
 \#2 & \textbf{0.00}  &  0.02  & 0.02 & \textbf{0.00}  &  0.07  & \textbf{0.00}  &  \textbf{0.00}  &  \textbf{0.00}  & \textbf{0.00}   \\ 
  \#3   & \textbf{0.00}  &  \textbf{0.16}  & 0.11 &  \textbf{0.00}  &  0.29  & 0.25   &  \textbf{0.00}  &  0.45  &\textbf{0.00}  \\ 
 \#4 & \textbf{0.30}  &  \textbf{1.27}  & \textbf{0.71} &  1.16  &  1.98  & 1.46   &  0.50  &  1.34  & 1.01 \\ 
   \#5 & \textbf{0.00}  &  \textbf{0.02}  & \textbf{0.00} &  \textbf{0.00}  &  0.09  & 0.05   &  \textbf{0.00}  &  0.45  & \textbf{0.00}  \\ 
 \#6 & \textbf{0.00}  &  0.07  & \textbf{0.00} &  \textbf{0.00}  &  0.16  & 0.11  &  \textbf{0.00}  &  \textbf{0.00}  & \textbf{0.00} \\ 
  \#7  & \textbf{0.00}  &  0.05  & \textbf{0.00} &  \textbf{0.00}  &  0.11  & \textbf{0.00}   & \textbf{0.00}  &  \textbf{0.00}  & \textbf{0.00}  \\


\bottomrule
\end{tabular}
\end{table*}

The unseen configuration selects one attack to be removed from training and to be the only one used for testing. This is to evaluate the capability of the model to generalize to novel attacks and to evaluate the challenges that each attack present to the network. The results for this configuration are seen in Table~\ref{unseen_attack}, and it is possible to observe that both DFS and adversarial approaches have difficulties in generalizing to unseen attacks. Especially attack \#4. The low performance of this attack can be explained by the high resolution of the replay attack device that increases the difficulty of the task.

\begin{table*}[h!]
 \caption{Evaluation of  three approaches in the setting of unseen attack in the ROSE-Youtu dataset. In this setting, the attack in the first column is excluded from the training and is the only one used for testing. APCER, BPCER and EER are displayed as \%. In bold is the best result per column.}
\label{unseen_attack}
\centering
\begin{tabular}{lccc|ccc|ccc}
\toprule
Attack     & \multicolumn{3}{c}{MT} & \multicolumn{3}{c}{Adversarial MT} & \multicolumn{3}{c}{DFS MT}  \\ 
\cmidrule{2-10}
      &APCER &BPCER &EER &APCER &BPCER &EER &APCER &BPCER &EER \tnote{c} \\
\midrule
  \#1 & \textbf{1.00}  &  \textbf{0.85}  & \textbf{0.95} &  \textbf{1.00}  &  6.90  & 2.87   &  \textbf{1.00}  &  6.90  & 2.30    \\ 
 \#2 & \textbf{0.00}  &  \textbf{0.49}  & \textbf{0.25}& \textbf{0.00}  &  1.18  & 0.27   &  \textbf{0.00}  &  4.90  & 0.67   \\ 
  \#3   & 3.09  &  3.41  & 3.27 &  \textbf{1.44}  &  \textbf{3.07}  & \textbf{2.61}   &  2.14  &  10.24  & 5.79  \\ 
 \#4 & 13.82  &  \textbf{3.88}  & \textbf{7.57} &  \textbf{12.66}  &  6.61  & 9.15   &  17.59  &  16.70  & 17.09\\ 
   \#5 & \textbf{0.00}  &  1.98  & \textbf{0.65} &  0.25  &  \textbf{0.58}  & \textbf{0.45}  &  0.50  &  4.90  & 1.49   \\ 
 \#6 & \textbf{0.00}  &  0.89  & 0.33 &  0.10  &  \textbf{0.58}  & \textbf{0.10}   &  \textbf{0.00}  &  4.01  & 1.00 \\ 
  \#7  & \textbf{0.35}  &  \textbf{3.63}  & \textbf{1.77} &  2.32  &  7.68  & 5.20   &  0.51  &  9.58  & 1.78   \\ 

\bottomrule
\end{tabular}
\end{table*}

The multi-task approach excelled at detecting both attacks and bonafide samples, achieving an equal-error rate better than any other approach. And thus, it was the approach used to compare with the state-of-the-art for the ROSE-YOUTU dataset. he methods compared report their results on similar train/test split, as specified on the database publication document~\cite{Haoliang2018}. Our results, as seen in Table~\ref{sota-table} are better than the state-of-the-art when we include background and slightly better when there is no background. Despite the good results, it is important to note that the methods presented in this document were not designed to be the best performing methods at cross-dataset configuration. Instead, they are intended to allow a relevant study regarding the presence of background for this specific dataset.

\begin{table}[h!]
 \caption{Comparison of the best proposed approaches, both with and without background, with the state-of-the-art. In bold is the best result per column.  }
\label{sota-table}
\centering
\begin{tabular}{lc}
\toprule
Method      & EER (\%) \\ 
\midrule
 CoALBP (YCBCR)~\cite{Haoliang2018}& 17.1 \\
 CoALBP (HSV)~\cite{Haoliang2018}& 16.4 \\  
 Color~\cite{Boulkenafet2016,Du2021} & 13.9\\
 De Spoofing~\cite{jourabloo2018face,Du2021} & 12.3 \\
 RCTR-all spaces~\cite{Du2021}& 10.7\\
 ResNet-18~\cite{he2016deep} & 9.3\\
 SE-ResNet18~\cite{hu2018squeeze} & 8.6\\
 AlexNet~\cite{Haoliang2018}& 8.0 \\ 
 DR-UDA (SE-ResNet18)~\cite{wang2021} & 8.0\\
 DR-UDA (ResNet-18)~\cite{wang2021} & 7.2\\
 3D-CNN~\cite{li2018}& 7.0 \\ 
 Blink-CNN~\cite{Hasan2019}& 4.6\\
 DRL-FAS~\cite{cai2020drl} & 1.8 \\
 
 \midrule
 Ours w/ Background & \textbf{0.2} \\ 
\bottomrule
\end{tabular}
\end{table}

\begin{figure}[!h]
\centering
\includegraphics[width = \linewidth]{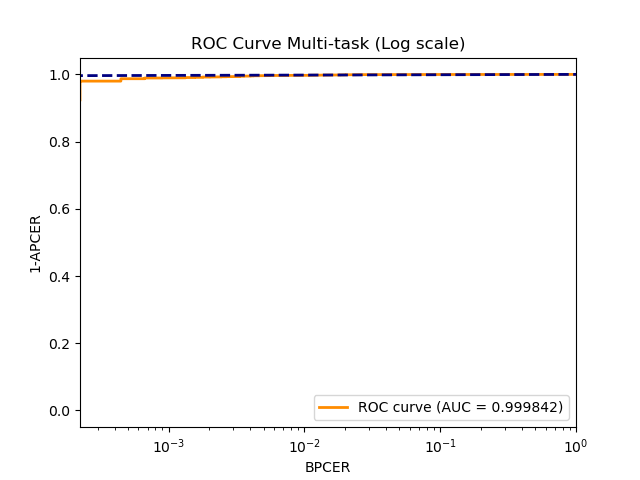}
\caption{Receiver operating characteristic curve for the multi-task model on the ROSE-Youtu dataset with background. X-axis is at log-scale. }
\label{roc_curve}
\end{figure}

In Figure~\ref{roc_curve} is depicted the ROC curve of the MT model with the x-axis displayed at the log-scale for a better visualization. The model is indeed nearly perfect at detecting the attacks and the bona-fide images in the ROSE-Youtu dataset.

\begin{figure}
    \centering
    \subfloat[Replay - B \label{1_back}]{%
       \includegraphics[height=2.1cm,width=1.8cm]{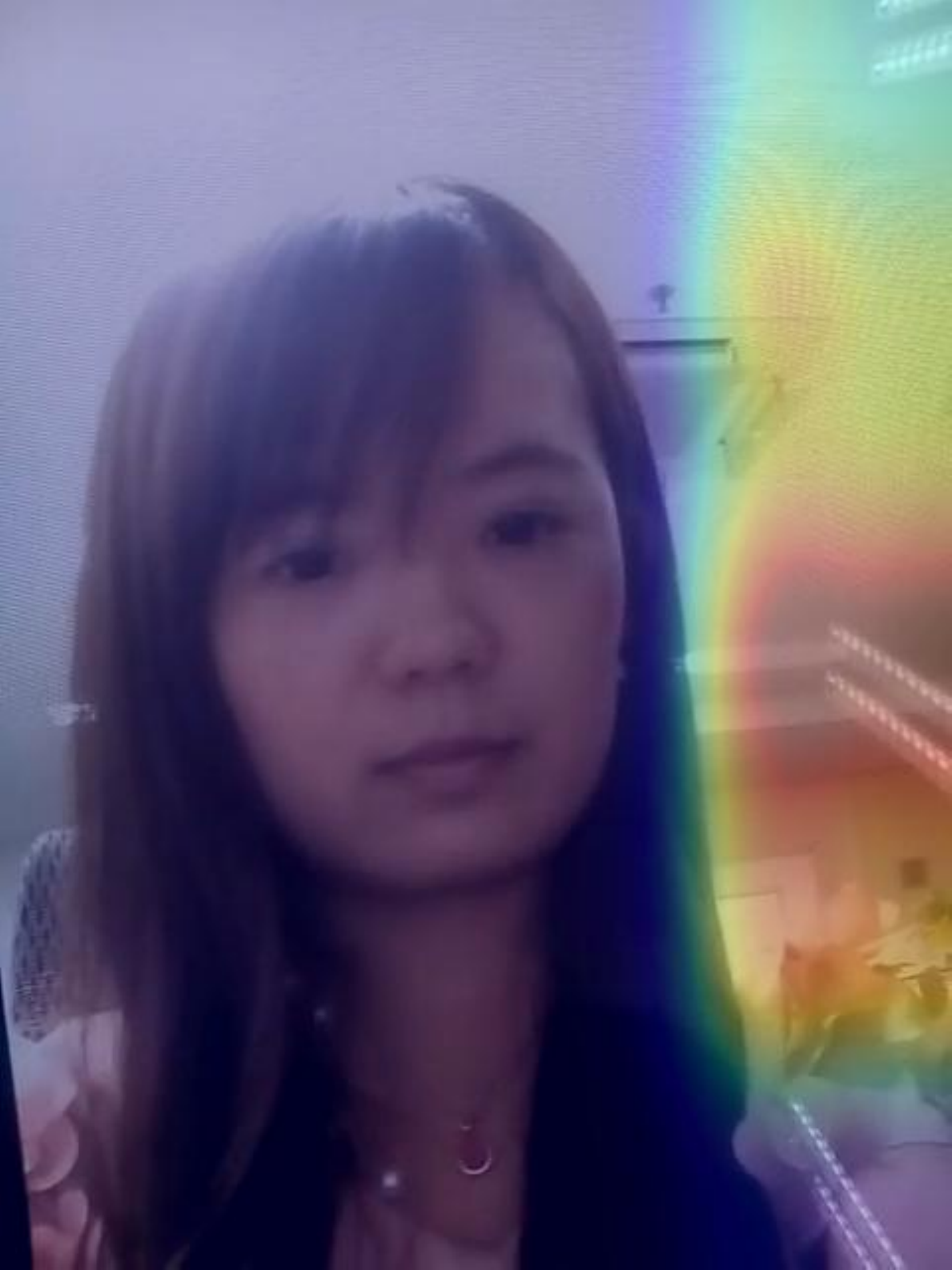}}
    ~
    \subfloat[Paper Mask - B\label{2_back}]{%
        \includegraphics[height=2.1cm,width=1.8cm]{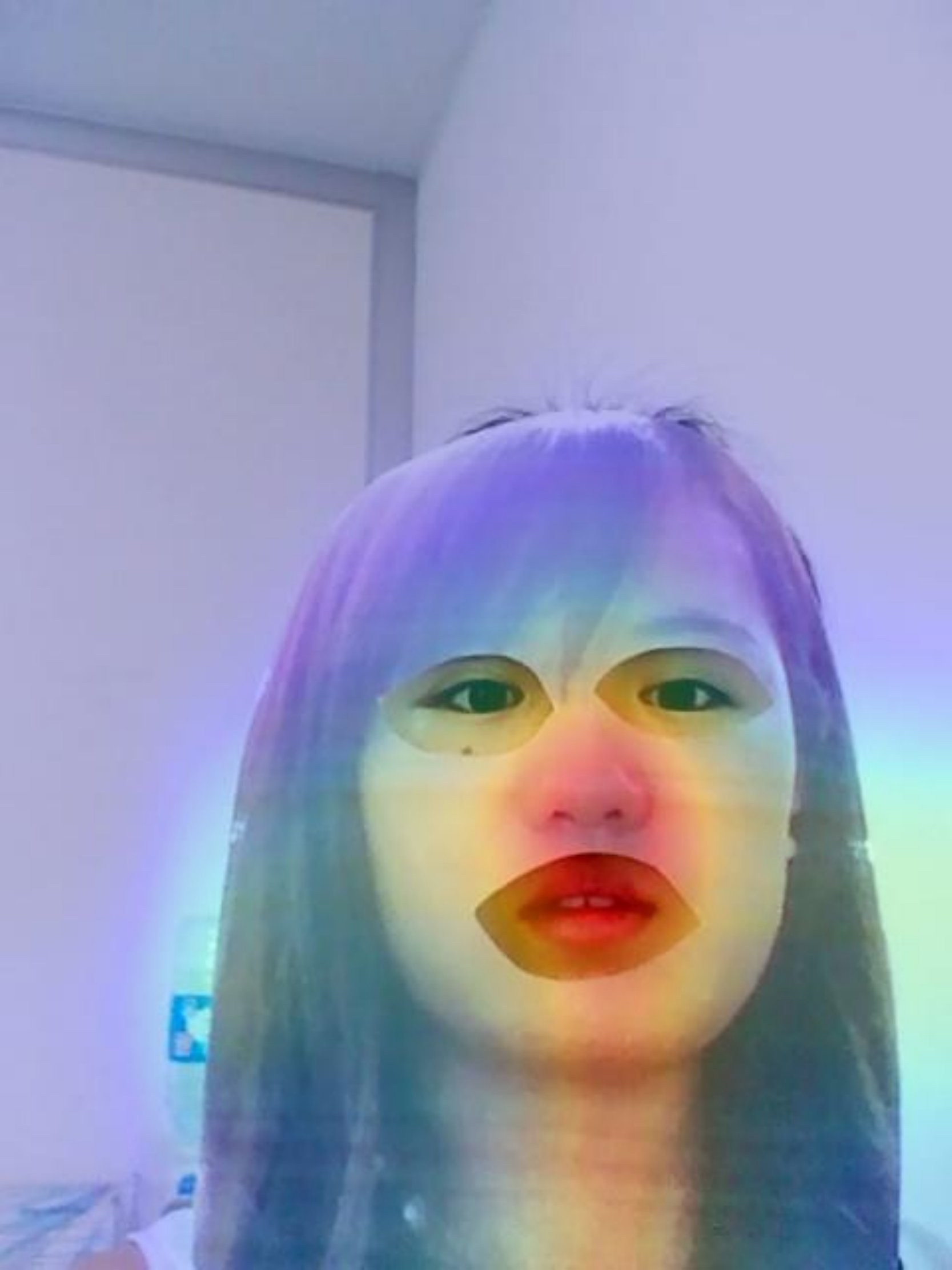}}
    ~
    \subfloat[Print - B\label{3_back}]{%
        \includegraphics[height=2.1cm,width=1.8cm]{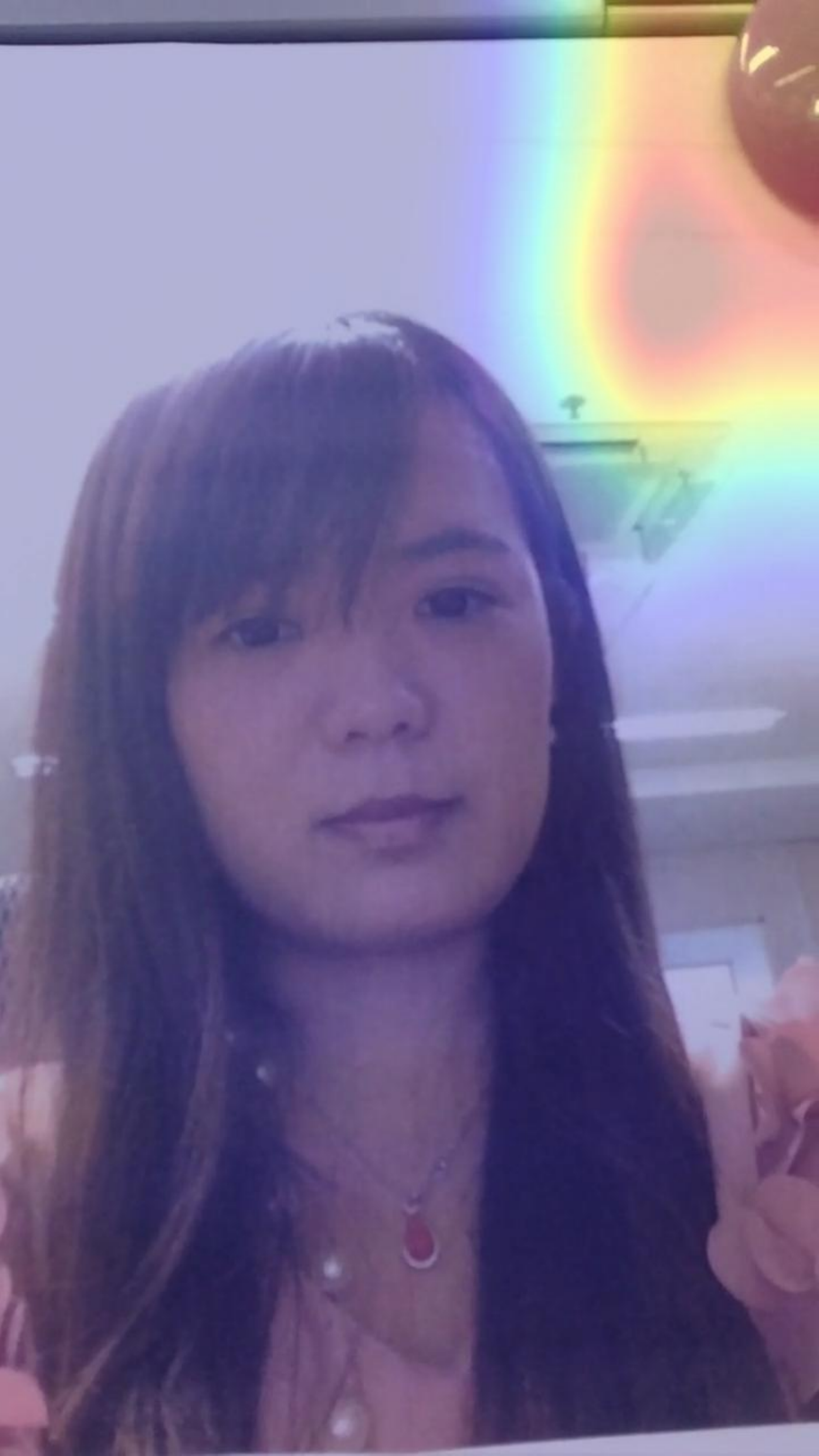}}
    \\
    \subfloat[Replay \label{1_crop}]{%
       \includegraphics[height=2.1cm,width=1.8cm]{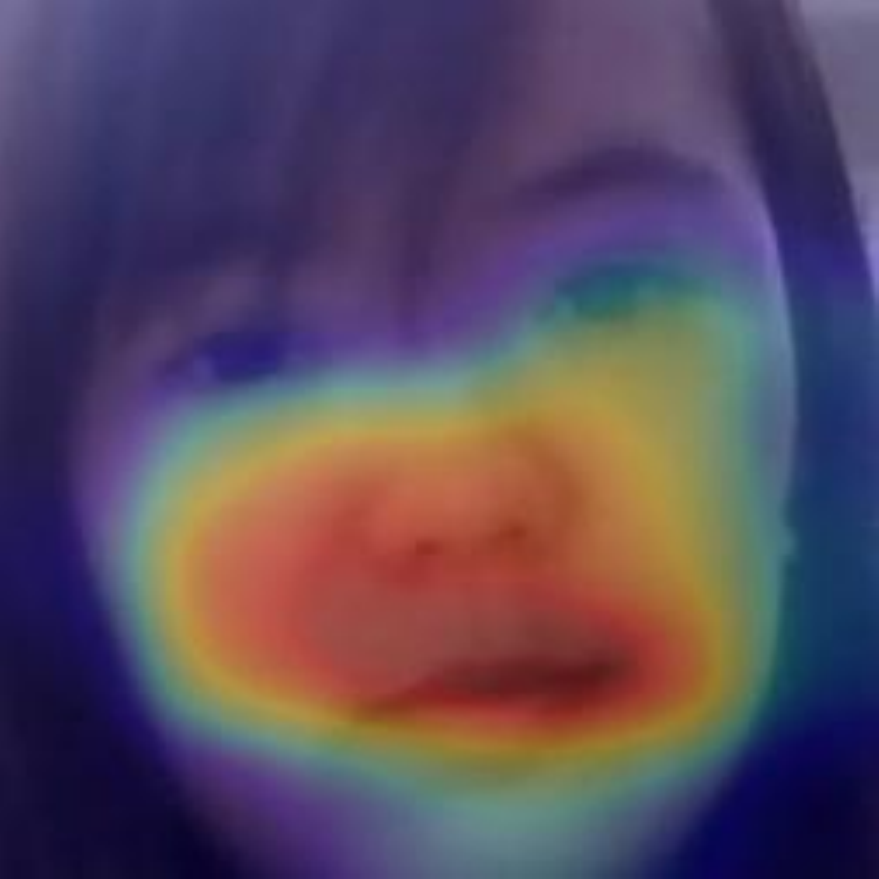}}
    ~
    \subfloat[Paper Mask \label{2_crop}]{%
        \includegraphics[height=2.1cm,width=1.8cm]{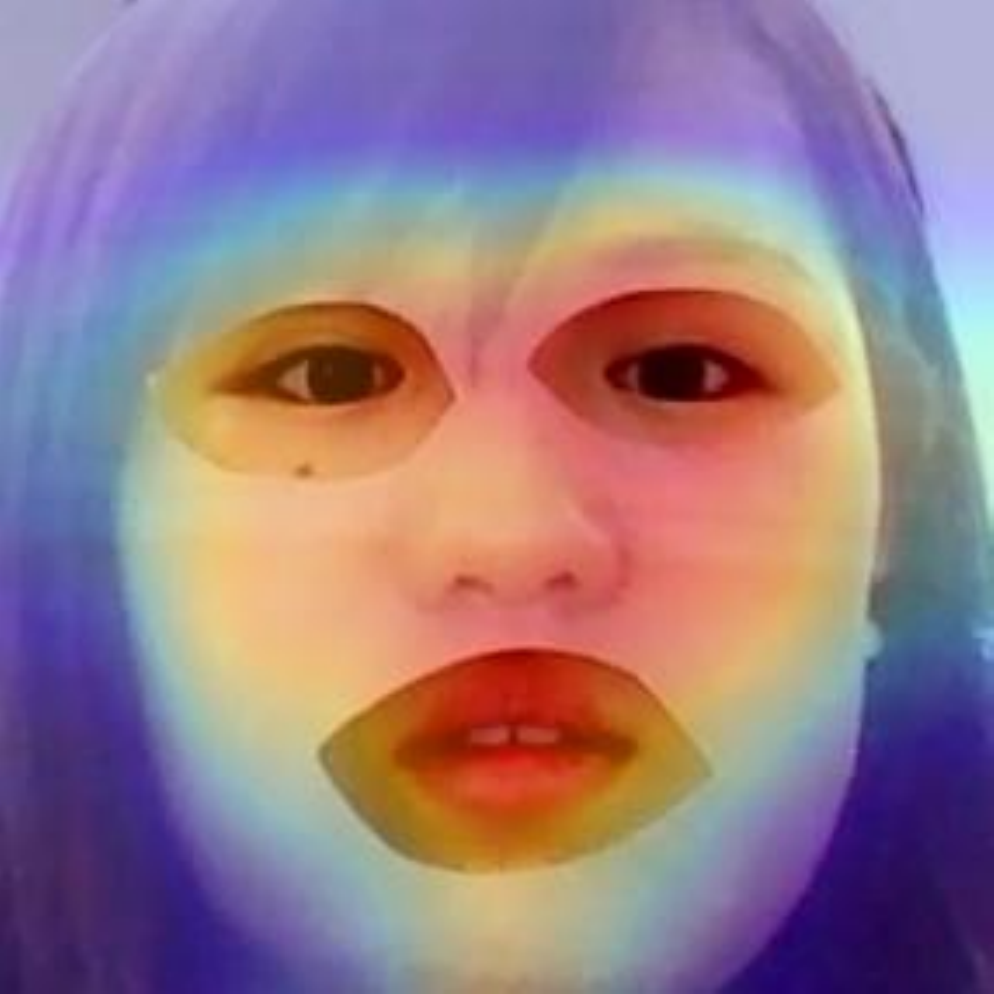}}
    ~
    \subfloat[Print \label{3_crop}]{%
        \includegraphics[height=2.1cm,width=1.8cm]{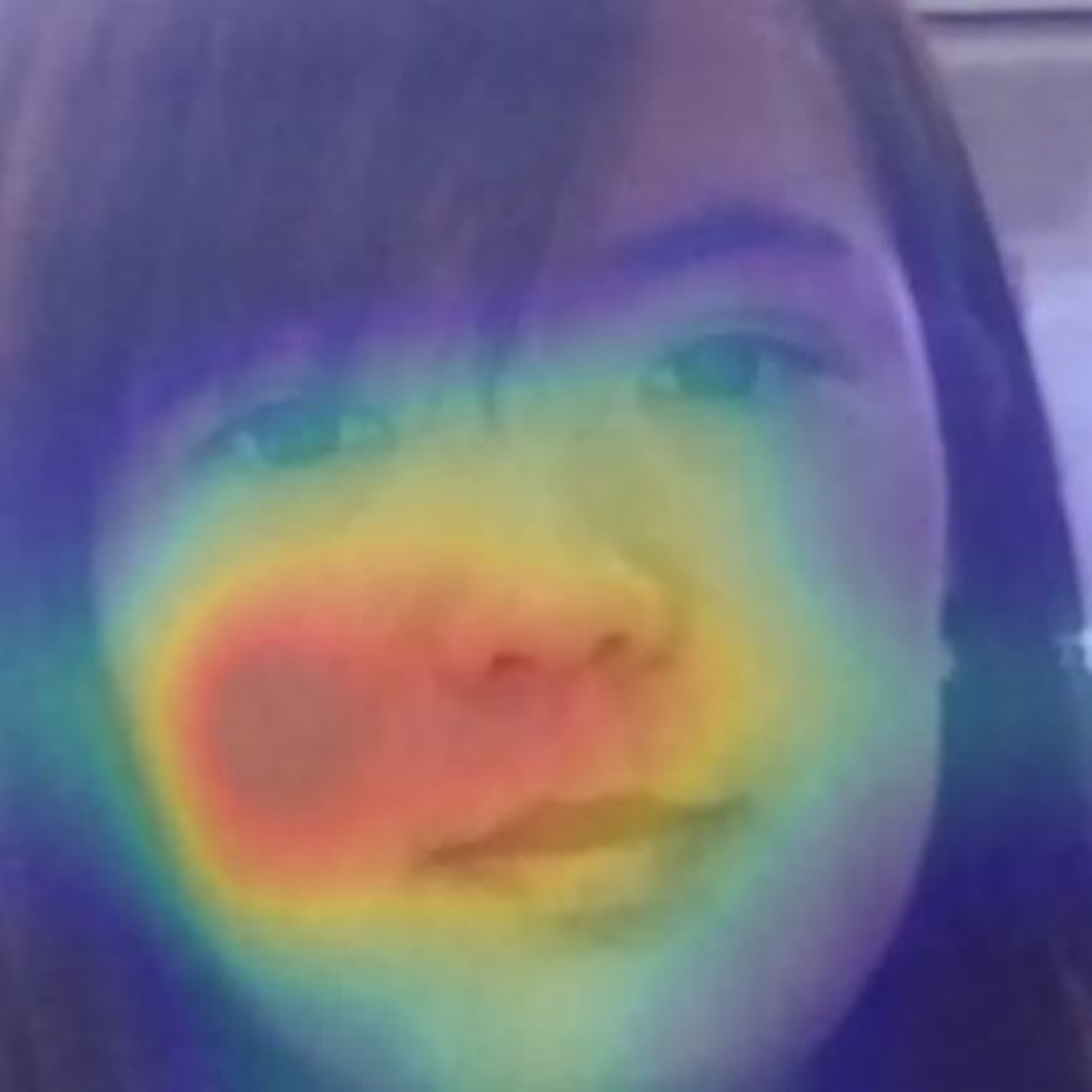}}
    \caption{Explanations generated with Grad-CAM++ for different attacks of subject \#23. -B indicates that the model was trained and tested on images with background. The (d) image was wrongly classified as genuine.
    \label{replay_attack} 
    }
\end{figure}

Finally, we produced explanations of our model for an example of each category of attacks. For the replay attack, we produced the explanations in Figures~\ref{1_back} and~\ref{1_crop}. In these figures, it is possible to observe that the models leveraged the presence of reflections in the attack image, whenever there is background. When the background is not present there are no cues to justify the decision of the model, which is in fact wrong. Figures~\ref{2_back} and~\ref{2_crop} shows the explanations for a paper mask attack, and as expected, the explanations do not rely on the background. Instead, the model directs its focus to the mask area for the final prediction. The area is similar on both versions of the model with and without background. Finally, the print attack explanations are seen in Figures~\ref{3_back} and~\ref{3_crop}. These figures shows that once more the model is capable of understanding the conditions of the image given and directs its focus to an important background artefact, the pin holding the image. And again, the version without background does not highlight any relevant cue that explains the prediction of the model. Hence, we further argue in favor of a better explanability factor in the models that include background.

\section{Conclusion}

This work explored how consistently the background impacts the performance of distinct methods for face presentation attack detection. The experiments corroborated the view that a face PAD model is capable of leveraging both background and face elements to make a correct prediction. 

Our approach surpassed the state-of-the-art results for the ROSE-YOUTU dataset by a large margin. The multi-task model leverages background artefacts to improve the detection of specific attacks. Moreover, we also present some alternative approaches, dynamic frame selection and adversarial training, that we believe were limited by the lack of a large database of face presentation attacks. Their results were consistent with the one from the multi-task model. 

We further contribute to improve the explainability of these models by analyzing the predictions. This analyze conducted with the Grad-CAM++ algorithm highlighted that models that include the background of the images can leverage the presence of certain artifacts. On the other hand, when the background is not present the generated explanations seem to be non-informative. Hence, due to their similarity with the human vision with regards to the areas used for the prediction, models that leverage the background provide more explanations for their predictions. 

\section*{Acknowledgments}
The authors would like to acknowledge the reviewers' comments that were crucial for the improvement of this work. This work was financed by National Funds through the Portuguese funding agency, FCT - Fundação para a Ciência e a Tecnologia within project UIDB/50014/2020, and within the PhD grant ``2021.06872.BD''.

\section*{Conflict of Interest}
The authors declare that there is no conflict of interest that could be perceived as prejudicing the impartiality of the research reported.

{\small
\bibliographystyle{ieee_fullname}
\bibliography{egbib}
}

\end{document}